\documentclass[letterpaper]{article} % DO NOT CHANGE THIS
\usepackage{aaai25}  % DO NOT CHANGE THIS
\usepackage{times}  % DO NOT CHANGE THIS
\usepackage{helvet}  % DO NOT CHANGE THIS
\usepackage{courier}  % DO NOT CHANGE THIS
\usepackage[hyphens]{url}  % DO NOT CHANGE THIS
\usepackage{graphicx} % DO NOT CHANGE THIS
\usepackage{natbib}  % DO NOT CHANGE THIS AND DO NOT ADD ANY OPTIONS TO IT
\usepackage{caption} % DO NOT CHANGE THIS AND DO NOT ADD ANY OPTIONS TO IT
\usepackage{listings,lstautogobble}
\usepackage{xcolor}
\definecolor{codegreen}{rgb}{0,0.5,0}
\definecolor{codeblue}{rgb}{0.25,0.5,0.5}
\definecolor{codegray}{rgb}{0.6,0.6,0.6}
\usepackage{amsfonts}

\newcommand{\tableref}[1]{Table~\ref{#1}} 

\newcommand{\figref}[1]{Fig.~\ref{#1}} 
\usepackage{subcaption}
\newtheorem{definition}{Definition}
\usepackage{amsmath}
 \usepackage{enumitem}
\usepackage{multirow}
\usepackage{makecell}
\usepackage{algorithm}
\usepackage{algorithmic}
\usepackage{fancyhdr}
\usepackage{newfloat}
\usepackage{listings}
\DeclareCaptionStyle{ruled}{labelfont=normalfont,labelsep=colon,strut=off} % DO NOT CHANGE THIS
\floatstyle{ruled}
\newfloat{listing}{tb}{lst}{}
\floatname{listing}{Listing}

\title{Learning to Balance: Diverse Normalization for \\ Cloth-Changing Person Re-Identification} 
\author{
	\parbox{\linewidth}{\centering
		Hongjun Wang\textsuperscript{\rm 1},
		Jiyuan Chen\textsuperscript{\rm 2},
		Zhengwei Yin\textsuperscript{\rm 1},	
		Xuan Song\textsuperscript{\rm 3},	
		Yinqiang Zheng\textsuperscript{\rm 1},	
	} \\
	{}
}
\affiliations{
	\textsuperscript{\rm 1} Department of Mechano-Informatics, The University of Tokyo;\\
	\textsuperscript{\rm 2} Department of Computing, The Hong Kong Polytechnic University;\\
	\textsuperscript{\rm 3} School of Artificial Intelligence, Jilin University; \\
}

\usepackage{bibentry}
\begin{document}

\maketitle

\thispagestyle{fancy} % IEEE模板在\maketitle后会自动声明\thispagestyle{plain}，
% 导致第一页什么都没有。所以得把plain更改为fancy
\lhead{} % 页眉左，需要东西的话就在{}内添加
\chead{} % 页眉中
\rhead{} % 页眉右
\lfoot{} % 页眉左
\cfoot{} % 页眉中
\rfoot{\thepage} %页眉右，\thepage 表示当前页码
\renewcommand{\headrulewidth}{0pt} %改为0pt即可去掉页眉下面的横线
\renewcommand{\footrulewidth}{0pt} %改为0pt即可去掉页脚上面的横线
\pagestyle{fancy}
\rfoot{\thepage}

\begin{abstract}
Cloth-Changing Person Re-Identification (CC-ReID) involves recognizing individuals in images regardless of clothing status. In this paper, we empirically and experimentally demonstrate that completely eliminating or fully retaining clothing features is detrimental to the task. Existing work, either relying on clothing labels, silhouettes, or other auxiliary data, fundamentally aim to balance the learning of clothing and identity features. However, we practically find that achieving this balance is challenging and nuanced.
In this study, we introduce a novel module called Diverse Norm, which expands personal features into orthogonal spaces and employs channel attention to separate clothing and identity features. A sample re-weighting optimization strategy is also introduced to 
guarantee the opposite optimization direction. Diverse Norm presents a simple yet effective approach that does not require additional data. Furthermore, Diverse Norm can be seamlessly integrated ResNet50 and significantly outperforms the state-of-the-art methods.
%\textcolor{magenta}{\textit{The code is released in our supplementary material.}}
\end{abstract}

\section{Introduction}
\label{sec:intro}

Person Re-identification (ReID), which aims to match a target person's image across different camera views, is highlighted  as a crucial task in the field of intelligent surveillance systems \cite{shi2020identity} and multi-object tracking \cite{ke2019multi}. In the early works of ReID, researchers normally take the assumption that people don't change their clothes in a short period of time. Therefore, methods developed during this stage mostly leverage the clothing information to identify people \cite{gu2020appearance,zhou2019omni,Zheng2015Scalable, gu2019TKP,Hou2021FC}. Generally, these methods can perform well in short-term datasets but will suffer from significant performance degradations when testing on a long-term one where people change their clothes frequently. To overcome such limitations, researchers has gradually turn their attention to  consider the change of clothes in model training and testing.

\begin{figure}[t]
	\centering
	\includegraphics[width=1\linewidth]{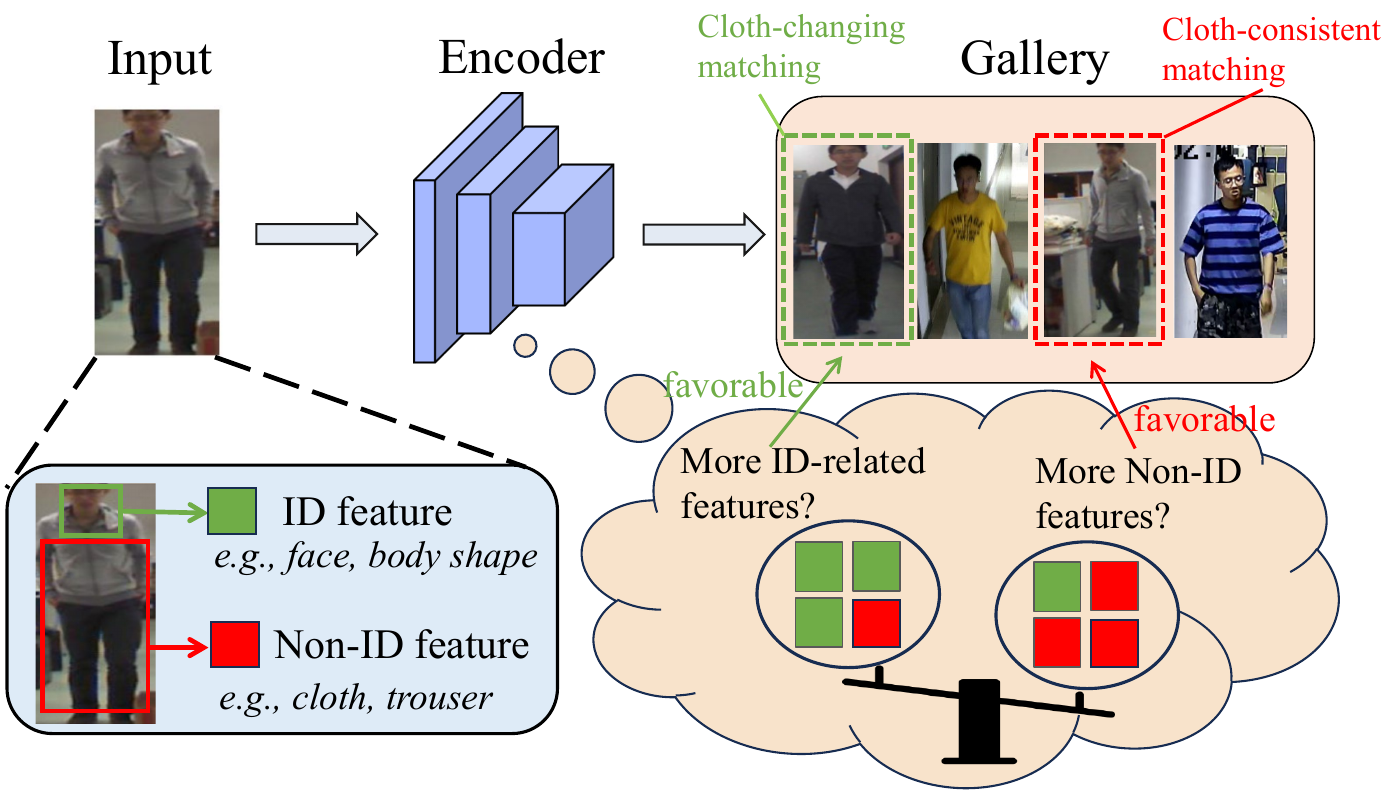}
	\caption{\textbf{A trade-off exists between maintaining clothing consistency and  clothing changes features in CC-ReID.} Traditionally, the model must acquire two distinct sets of features: one focused on clothing features (such as garments and trousers) and another on clothing-irrelevant attributes (like face and body shape) to effectively handle gallery with same person involving cloth consistency and cloth changing, respectively. Finally, either clothes or ID-invariant features will be used to match galleries  images of the same person with different clothing status. }\label{fig:motivation}
	% \vspace{-0.5cm}
\end{figure}

\begin{figure}[t]
	\centering
	\includegraphics[width=0.9\linewidth]{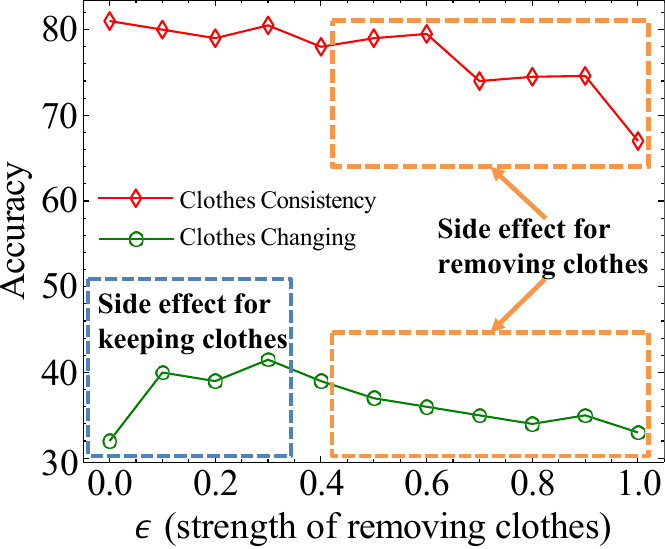}
	\caption{The relation between accuracy and strength of removing clothes features in CAL \cite{gu2022clothes}.  We found that completely removing clothing features is not beneficial; instead, it can simultaneously disrupt scenarios where clothing is not changed. However, if too many clothes features are retained, it will not be good for the clothes changing scenes. }
	\label{fig:tradeoff}
\end{figure}

With years of developments, the mainstream solutions of cloth-changing person re-identification (CC-ReID) are to either incorporate more ID features from other modalities (\textit{e.g.}, contour sketch \cite{yang2019person}, body shape \cite{qian2020long}, hairstyle \cite{wan2020person,yu2020cocas} and 3D shape \cite{chen2021learning}) and encourage the model to learn from them, or use hand-crafted clothes labels \cite{gu2022clothes,cui2023dcr,yang2023good} to force the model pay less attention to the clothes appearance in CC-ReID task. However, as illustrated in \figref{fig:motivation}, regardless of which path is taken, we argue that all these works are actually trying to find a best trade-off point for encoding ID features and clothing features into the high-level representation of the model. To further validate this conjecture, we conducted grid searching experiments on CAL \cite{gu2022clothes} and illustrated the result in \figref{fig:tradeoff}. Specifically, we adjusted the intensity of the adversarial loss, denoted as $\epsilon$. A larger $\epsilon$ indicates the removal of more clothing features. We found that completely removing clothing features is detrimental; it can also negatively impact scenarios where clothing remains unchanged. Conversely, retaining too many clothing features is detrimental to scenarios involving clothing changes.
The reason for this phenomenon is quite intuitive. As illustrated in \figref{fig:motivation}, both the query and gallery datasets in CC-ReID contain scenarios of both clothes-changing and clothes-consistency person. Therefore, the optimal model represents the best equilibrium point between these two scenarios.

When viewing the CC-ReID problem from the ``trade-off" perspective, we find that the previous efforts of using multi-modality or clothes labels are all for separating the learning of clothing and ID features so as to balance their proportion. However, we notice that it is unnecessary with so much effort because the learnings of clothing and ID features are naturally separated to each other (\textit{i.e.}, a feature cannot simultaneously belongs to both clothing-relevant and ID-invariant feature groups).  In \figref{fig:tradeoff}, we further notice that ``trade-off" searching in existing works is actually an ill-posed optimization problem with two optimization directions fighting with each other in the same model. Therefore, we can actually separating such learning using only RGB modality and free from acquiring clothes labels with only a simple whitening operation. Inspired by the ideas from ensemble learning,  this paper explicitly constructs two separate branches with each one focusing on one optimization target, and the diversity between them is ensured by a whitening operation \cite{huang2018decorrelated} and  channel attention \cite{hu2018squeeze}. Moreover, sampling re-weighting strategy \cite{kim2022learning} is also used to serve as a regularization to assign opposite sample weights to let different branches focus on specific input samples. For instance, the back-view sample is heavily weighted towards the clothing branch, while the face-forward sample is prioritized for the ID branch.  

\noindent\textbf{Contributions:} 1) For the first time, we points out that a potential bottleneck in existing work may arise from  ill-posed optimization issue and a tendency to become trapped in local minima. 2) To address this issue, we present Diverse Norm for CC-ReID, which is designed to create orthogonal features from the same person embeddings, allowing for meeting the requirement of short-term and long-term ReID. 3)  Our provide compelling evidence that Diverse Norm surpasses other state-of-the-art methods, including both multi-modality-based  and  clothing labels-based approaches.

% The sampling re-weighting strategy is widely be used in debiasing modeling 
%\cite{lee2021learning,lee2023revisiting}
%theoretically ensures distinct optimization directions for each branch and prevents them from learning the same concept.

%To sum up, we present a novel whitening method called as Diverse Norm, designed to extract two distinct features from the same person embedding. These features, orthogonal in nature, undergo separate optimization during training and are later fused in testing based on their distance to the gallery image. This fusion allows for a dynamic weighting of consistency and clothing changes feature, as shown in \figref{fig:motivation}.
%Diverse Norm enhances the performance of a pure ResNet50 to surpass all SOTA methods in CC-ReID. 

%\vspace{-1pt}

\begin{figure*}[t]
	\centering
	\includegraphics[width=1\linewidth]{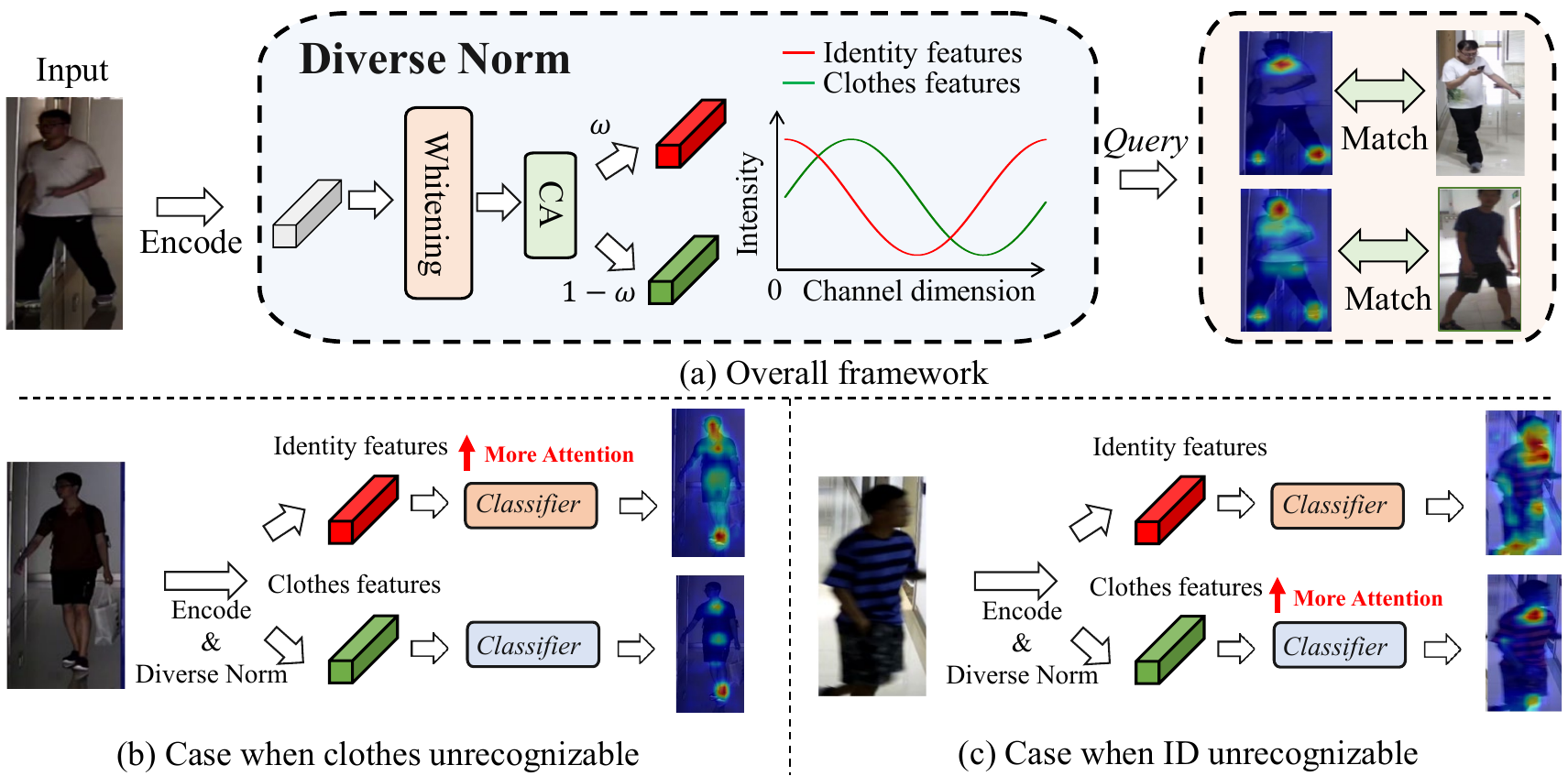}
	\caption{\textbf{Architecture Overview  of Our Method.} In Fig. (a), we first apply whitening to the features extracted by the backbone network and then utilize channel attention to separate clothing and identity features. To effectively distinguish between clothes and identity features, we constructed two classifiers and employed sample reweighting to achieve the concept selection. Specifically, as shown in Fig. (b), when an individual is in a dimly lit area wearing dark clothing, making recognition based solely on clothing difficult, the identity branch increases its weight. Conversely, as depicted in Fig. (c), when a person is moving quickly, causing facial motion blur, but wearing distinct plaid clothing, the apparel branch increases its weight. 
	  } \label{fig:framework}
\end{figure*}

\section{Related Work}\label{sec:related_work}
\noindent\textbf{Long-Term Person Re-Identification.}
%Early approaches to person Re-IDentification (ReID) focused primarily on either feature extraction methods \cite{wang2007shape, ma2012bicov, corvee2010person, farenzena2010person} or metric learning techniques \cite{koestinger2012large, li2013learning, chen2016similarity, mignon2012pcca}. In contrast, recent methods have benefited significantly from the advancements in Convolutional Neural Network (CNN) architectures, which enable end-to-end learning \cite{Chen_2017_CVPR_deep, Li_2018_CVPR_deep, Li_2017_CVPR_deep, Wang_2016_CVPR_deep, Xu_2018_CVPR_reid, Xiao_2016_CVPR_deep}. Currently, the ReID problem has been addressed by training with an identification loss \cite{Xiao_2017_CVPR}, contrastive loss \cite{varior2016gated, Varior2016} or triplet loss \cite{Cheng_2016_CVPR_reid, Chang_2018_CVPR_reid}.
%However, existing clothing-consistent based models struggle in CC-ReID settings due to the drastic changes in clothing appearances over time. 
The primary focus in clothes-changing re-identification (re-id) research is to extract features from images that are not influenced by clothing.  Within this field, two main approaches have emerged:
1) A branch of
research  attempts
to use disentangled representation learning \cite{Zhang2019Gait, DGNet,chan2023learning,gu2022clothes} to separate appearance from structural information in RGB images. This approach treats structural information as features that are independent of clothing. 2) Another branch of research aim to combine multi-modality data, such as,  skeletons \cite{Yang2016Learning,Ariyanto2012Marionette}, silhouettes \cite{Chao2019Gaitset,Han2006Individual}, 3D information \cite{liu2023learning,Chen2021Learning3D} to exert the structural information. 
However, we found that achieving this balance is fragile. Therefore, this paper attempts to optimize the CC-ReID problem using a decoupled approach.
%However, existing methods are both time-consuming and require additional equipment. Additionally, they constrain the model to learn a complex system where the balance between clothing-related attributes and individual characteristics is not well-defined, making it difficult to optimize. This paper introduces Diverse Norm, generating a orthogonal space for clothing-related and clothes-independent features.

\noindent\textbf{Whitening and Orthogonality.}
%Whitening, a linear transformation used in data science, reshapes the covariance matrix of input vectors into an identity matrix. In deep learning, batch normalization \cite{ioffe2015batch} is a popular technique that standardizes data but does not address decorrelation. Some earlier approaches attempted whitening through periodic estimation of the whitening matrix \cite{desjardins2015natural,luo2017learning}, resulting in training instability. Others introduced a decorrelation loss for whitening \cite{cogswell2015reducing}.
%Recent advances in ZCA whitening introduced differentiable whitening modules based on Singular Value Decomposition (SVD) \cite{huang2018decorrelated, huang2019iterative}, and Cholesky whitening \cite{siarohin2018whitening}. 
In deep learning, orthogonality constraints have been applied to address vanishing or exploding gradients, particularly in Recurrent Neural Networks (RNNs) \cite{vorontsov2017orthogonality, mhammedi2017efficient, wisdom2016full}. These constraints have been extended to various neural network types, including non-RNNs \cite{harandi2016generalized, huang2018orthogonal,lezcano2019cheap,lezama2018ole}. Some methods use specialized loss functions to enforce orthogonality \cite{lezama2018ole}. In contrast, CW optimizes the orthogonal matrix using Cayley-transform-based curvilinear search algorithms \cite{wen2013feasible}, with the unique goal of aligning orthogonal matrix columns with specific concepts, distinguishing it from prior methods. Our Diverse Norm module incorporates whitening techniques from IterNorm \cite{huang2019iterative} to separate appearance from structural information in RGB images.

\section{Methodology}

\noindent\textbf{Diversity Normalization.}
As mentioned in \figref{fig:tradeoff}, achieving a balance between clothing and identity features is challenging, and even slight missteps can degrade the performance of one aspect. To address this issue, we introduce an expand operation, aiming to learn robust representations of both clothing and identity features during training. Specifically, we introduce whitening \cite{huang2018decorrelated} to disentangle the conception \cite{chen2020concept}, combined with channel attention (CA) \cite{hu2018squeeze} for clothes and identity features selection. Formally, whitening is defined as:
\begin{definition}
	Whitening is designed to process latent features in a latent space through decorrelation, standardization (whitening), and orthogonal transformation. The whitening transformation can be represented as:
	\begin{equation}
		\psi(\mathbf{Z}) = \mathbf{W}(\mathbf{Z} - \mathbf{\mu}),
	\end{equation}
	where $\mathbf{\mu}$ is the sample mean and $\mathbf{W}$ is the whitening matrix satisfying $\mathbf{W}^\top \mathbf{W} = \mathbf{\Sigma}^{-1}$, with $\mathbf{\Sigma}$ is the covariance matrix.
\end{definition}
%As clothing and identity features do not spatially overlap, we use CA to separate them. 
Given the human features $h$ generated by the backbone network, we then integrate CA into our architecture to separate clothing and identity features.  Specifically, the channel weights in the CA block are calculated as:
$$
\omega=\sigma\left(Wg(\psi(h))\right) \in (0, 1),
$$
where $g$ is the channel-wise Global Average Pooling (GAP), $W$ is fully-connected layer, and $\sigma$ is the Sigmoid function. As clothing and identity features do not spatially overlap, we use $\omega$ and $1 - \omega$ to decompose $\hat{h}$ into $h_{ID}$ and $h_{C}$:
\[
h_{id} = \psi(h) \cdot \omega, \quad h_{c} = \psi(h) \cdot (1 - \omega)
\]
where $h_{id}$ and $h_c$ represent identity features and clothing features, respectively.

\noindent\textbf{Sample Re-weighting for Conception Separation.}
Obviously, the whitening and CA only guarantee the orthogonality of feature pairs and feature selection, which necessitates further strategies to ensure the separation between clothing-relevant and clothing-irrelevant features.
As shown in \figref{fig:framework}, we observe that among training samples, some are identifiable by clothing, while others are recognizable by identity features. Based on this observation, we employ a  re-weighting strategy \cite{nam2020learning,creager2021environment} to amplify this distinction, encouraging classifiers to be biased towards certain types of samples. Formally, given the classifier output $\tilde{y}(h_{id})$ and $\tilde{y}(h_c)$, the relative difficulty score  $\mathcal{W}(h_{id})$ and $\mathcal{W}(h_c)$ are defined as follows:
\begin{align*}
	\mathcal{W}(h_{id}) \equiv 1, \ \mathcal{W}(h_c)=\frac{2\mathcal{L}_{ID}\left(\tilde{y}(h_c), y\right)}{\mathcal{L}_{ID}\left(\tilde{y}(h_{id}), y\right) + \mathcal{L}_{ID}\left(\tilde{y}(h_c), y\right)},
\end{align*}
where $\mathcal{W}(h_{id})$ serves as an anchor \cite{creager2021environment}, and always equals 1,  $\mathcal{W}(h_c)$ is the ratio of loss between $h_{id}$ and $h_c$, and  $\mathcal{L}_{ID}$ indicates cross entropy loss for classification. We draw upon \cite{creager2021environment} to define the anchor, which represents a \textit{biased} branch. In \ref{fig:tradeoff}, the baseline is assumed to be biased towards the branch that predominantly captures clothing features. This assumption is reasonable, considering that clothing can account for up to 50\% of the visual content in an image. Hence,  the relative difficulty score $\mathcal{W}(h_c)$ is employed to guide the identity classifier during its learning process.

\noindent\textbf{Similarity between Query and Gallery Data.}
Let \( q_{id} \), \( q_c \), \( g_{id} \), and \( g_c \) represent the identity and clothing features of the query and gallery images, respectively. We employ cosine distance to measure the similarity between the feature  of the query and gallery images and then aggregate these similarities. Our experiments indicate that this approach significantly enhances model performance. We also compared the effects of combining \( q_{id} + q_c \) and \( g_{id} + g_c \) in \figref{fig:top1}. We found that the primary reason for the traditional baseline's failure is the overshadowing of ID features, which account for a much smaller proportion compared to clothing features.

%For example, at the onset of training, we assign $\mathcal{W}(h_{id})$ to branch one and $\mathcal{W}(h_c)$ to branch two. Subsequently, branch with $\mathcal{W}(h_{id})$  naturally inclines towards capturing clothing features. Consequently, at each gradient update step, branch with  $\mathcal{W}(h_c)$ will concentrate more on clothing-changing samples and acquire ID-invariant  features. 

\begin{table*}[t]
	\centering
	\setlength{\abovecaptionskip}{-2pt}
	\setlength{\belowcaptionskip}{-4pt}
	\renewcommand{\arraystretch}{1.3}
	\caption{ \textbf{Comparison with state-of-the-art methods on LTCC, and PRCC.}  The results indicated by an underline were copied from the original papers. A backslash indicates that a result is not available.  The Extra data indicates, such as,  contour sketches, silhouettes, human poses, and 3D shape information.}
	%	\vspace{-1pt}
	% \setlength{\belowcaptionskip}{-0.2cm}
	% \small
	\begin{sc}
		\resizebox{\textwidth}{!}{
			\begin{tabular}{l|c|c|ccc|ccc|ccc|ccc}
				\hline \multirow{3}*{Methods} &\multirowcell{3}{Cloth\\labels}&\multirowcell{3}{Extra\\data} &\multicolumn{6}{c}{LTCC}  &\multicolumn{6}{c}{PRCC}\\
				\cline{4 - 15}
				&&&\multicolumn{3}{c}{General} &\multicolumn{3}{|c}{CC} &\multicolumn{3}{|c}{Same Clothes} &\multicolumn{3}{|c}{CC}  \\
				\cline { 4 - 15 } && & mAP & Rank1 & Rank5 & mAP & Rank1 & Rank5 & mAP & Rank1 & Rank5 & mAP & Rank1 & Rank5\\
				\hline LTCC  & &$\checkmark$& $\underline{34.31}$ & $\underline{71.39}$ & $\backslash$ & $\underline{12.40}$ & $\underline{26.15}$ & $\backslash$ &89.37&96.08&98.24&28.89&30.45&38.98\\
				3APF  & &$\checkmark$& 21.37 & 54.41 & 70.23 & 9.72 & 22.43 & 38.76 &95.02&98.04&99.69&41.63&43.95&54.58\\
				FSAM  & &$\checkmark$& $\underline{35.4}$ & $\underline{73.2}$ & $\backslash$ & $\underline{16.2}$ & $\underline{38.5}$ & $\backslash$ &—&98.8&100.0&—&54.5&86.4 \\
				CVSL  &&$\checkmark$&  41.9 & \bf 76.4 &$\backslash$  & 21.3 & 44.5 &$\backslash$ & 99.1 & 97.5&$\backslash$ & 56.9 & 57.5&$\backslash$ \\
				3DSL  & $\checkmark$ &$\checkmark$& $\backslash$ & $\backslash$ & $\backslash$ & $\underline{14.8}$ & $\underline{31.2}$ & $\backslash$ &—& $\backslash$ & $\backslash$ &—&51.3&86.5\\
				CASE  & $\checkmark$ && 21.10 & 52.70 & 70.59 & 9.86 & 22.96 & 38.52 &96.18&98.89&99.41&36.68&42.90&56.34\\
				FD-GAN  &$\checkmark$&& 36.89&73.43&80.12&15.36&32.91&46.68&99.69 & 99.97 &100.00& 58.57& 58.34& 63.82\\ 
				AIM &$\checkmark$&  &41.1 &   76.3 & $\backslash$&19.1 &40.6 & $\backslash$ &  \bf 99.9 &  \bf 100 & $\backslash$& 58.3  & 57.9 & $\backslash$ \\
				CCFA  &$\checkmark$& & \bf 42.5 & 75.8  &$\backslash$ & \bf 22.1 &  \bf 45.3 &$\backslash$  & 98.7 & 99.6 &$\backslash$  &  \bf 58.4 &  \bf 61.2 &$\backslash$  \\ 
				\hline 
				ReIDCaps  & && 15.81 & 43.03 & 58.61 & 6.20 & 12.82 & 23.71 &96.62&99.54&99.61&44.81&48.00&54.61\\
				BoT  & && 27.80 & 64.33 & 76.91 & 10.70 & 28.12 & 42.12 &98.41&99.30&99.82&45.31&44.92&52.51 \\
				PCB  & && 30.39 & 64.50 & 76.27 & 11.24 & 21.27 & 38.78 &97.01&99.02&99.90&37.56&36.49&43.75 \\
				Part-aligned  & && 21.33 & 53.18 & 69.61 & 9.58 & 21.41 & 38.50 &95.63&98.70&99.82&39.33&42.24&51.51  \\
				MGN  & && 33.85 & 66.32 & 74.95 & 13.69 & 27.30 & 41.34 &98.77&99.92 &99.97&53.12&55.21&62.09 \\
				SCPNet  & && 18.02 & 48.07 & 67.34 & 7.38 & 15.82 & 31.63 &93.09&97.68&99.10&23.48&22.92&35.39\\
				ISGAN  &  && 33.76 & 68.56 & 77.69 & $\underline{13.03}$ & 29.08 & 42.86 &99.65&99.90&100.00&55.97&55.38&61.84\\
				DGNet  & && 32.18 & 64.71 & 78.30 & 10.80 & 21.70 & 34.69 &97.42&99.15&99.85&48.17&49.08&63.53\\
				\hline
				ResNet50  &&&  29.41  & 65.57& 73.80 &11.05  &28.16 & 42.13 & 98.44 & 97.90 & 99.81 & 43.36 & 45.67 & 52.32\\
				+Diverse Norm  & && \textcolor{red}{\textbf{46.48}}& \textcolor{red}{\textbf{78.92}}&\textcolor{red}{\textbf{82.45}}&\textcolor{red}{\textbf{31.88}}& \textcolor{red}{\textbf{63.30}}&\textcolor{red}{ \textbf{71.97}}&  \textcolor{red}{\textbf{99.89} }&\textcolor{red}{\textbf{99.94}}&\textcolor{red}{\textbf{100.00}}& \textcolor{red}{\textbf{58.87}}& \textcolor{red}{\textbf{59.74}}& \textcolor{red}{\textbf{64.42}} \\
				\hline
				TransReID   &&& 33.08 & 63.74 & 74.00  & 15.17 & 26.88 & 43.19 &   99.89 & 99.99 & 100.00 & 43.08 & 48.96 & 51.22 \\ 
				+Diverse Norm  & && \textcolor{red}{\bf 47.08} & \textcolor{red}{\bf 77.75} & \textcolor{red}{\bf 83.99} & \textcolor{red}{\bf 33.38} & \textcolor{red}{\bf 60.70} & \textcolor{red}{\bf 72.33} & \textcolor{red}{\bf 99.89} & \textcolor{red}{\bf 99.91} & \textcolor{red}{\bf 100.00} & \textcolor{red}{\bf 59.28} & \textcolor{red}{\bf 61.35} & \textcolor{red}{\bf 67.98}\\
				\hline

		\end{tabular}}
			\vspace{-5pt}
	\end{sc}
	\label{tab:res}
\end{table*}

\begin{table}
	\centering
	\caption{Comparison with state-of-the-arts on VC-Clothes}\label{tab:result1}
	\renewcommand\arraystretch{1.3}
	\begin{sc}
		\resizebox{0.48\textwidth}{!}{
			\begin{tabular}{l| c c| c c| c c}
				\hline
				\multirow{3}*{Method}   &\multicolumn{2}{c}{General} &\multicolumn{2}{|c}{SC}
				&\multicolumn{2}{|c}{CC}\\
				&\multicolumn{2}{c}{(all cams)} &\multicolumn{2}{|c}{(cam2\&cam3)}
				&\multicolumn{2}{|c}{(cam3\&cam4)}\\
				\cline{2-7}
				&top-1 &mAP &top-1 &mAP &top-1 &mAP \\
				\hline
				MDLA~          &88.9  &76.8  &94.3  &93.9  &59.2  &60.8 \\
				PCB~        &87.7  &74.6  &94.7  &94.3  &62.0  &62.2 \\
				\hline
				Part-aligned~                    &90.5  &79.7  &93.9  &93.4  &69.4  &67.3 \\
				FSAM~ &-    &-      &94.7  &94.8  &78.6  &78.9 \\	
				3DSL~  &-    &-      &-     &-     &79.9  &81.2 \\		
				CAL~ &92.9 &87.2 &95.1 &95.3 &81.4 &81.7 \\
				\hline
				ResNet50                       &88.3  &79.2   &94.1  &94.3  &67.3  &67.9\\
				+Diverse Norm & \textcolor{red}{\bf 93.1} &\textcolor{red}{\bf  88.2} & \textcolor{red}{\bf 96.6 }& \textcolor{red}{\bf 95.8 }&\textcolor{red}{\bf  85.9} & \textcolor{red}{\bf 83.2 }\\
				
				\hline
		\end{tabular}}
	\end{sc}
\end{table}

\begin{table}[t]
	\centering
	\caption{Comparison on LaST and DeepChange.}\label{tab:result2}
	\renewcommand\arraystretch{1.3}
	\begin{sc}
		\resizebox{0.4\textwidth}{!}{
			\begin{tabular}{l| c c| c c}
				\hline
				\multirow{2}*{Method}   &\multicolumn{2}{c}{LaST} &\multicolumn{2}{|c}{DeepChange}\\
				\cline{2-5}
				&top-1 &mAP &top-1 &mAP  \\
				\hline
				OSNet~             &63.8  &20.9   &39.7  &10.3 \\
				ReIDCaps~&-     &-      &39.5  &11.3 \\
				BoT~                            &68.3  &25.3   &47.5  &13.0 \\
				mAPLoss~            &69.9  &27.6   &-     &- \\
				CAL  & 73.7 &28.8 &54.0 &19.0 \\
				\hline
				ResNet50  &69.4  &25.6   &50.6  &15.9 \\
				+Diverse Norm & \textcolor{red}{\bf 74.4} & \textcolor{red}{\bf 29.7} &  \textcolor{red}{\bf 56.9} & \textcolor{red}{\bf  19.4 }\\ 
				\hline
		\end{tabular}}
	\end{sc}
\end{table}

\section{Experiment}
In this paper, our experiment firstly involves two key CC-ReID datasets, PRCC~\cite{Yang2019PRCC} and LTCC~\cite{Qian2020LTCC}, which are essential for the current state of the arts's comparison. We here aim to prove that the the clothes labels for the CC-ReID is not necessary if we explicitly disentangled the clothes and identical features. 
PRCC dataset has 22,889 training and 10,800 test images, with test samples from A forming the gallery set, and B and C forming the query set for different matching scenarios \cite{Yang2019PRCC}.
LTCC, similar to PRCC, comprises 9,576 training images and 7,050 test images. The dataset is divided for same-clothes and cross-clothes matching, with 493 randomly selected query images from various cameras and outfits \cite{Qian2020LTCC}.   In LTCC, clothing information is readily available, while in PRCC, the clothing labels can be inferred conveniently from the camera IDs, which are used in 
3DSL \cite{Chen2021Learning3D}, CASE \cite{li2021learning}, FD-GAN \cite{chan2023learning}, CAL \cite{gu2022clothes}, CVSL \cite{nguyen2024contrastive}.

During the evaluation phase, ResNet50 with Diverse Norm calculates the average similarity of branches $h_{id}$ and $h_c$ between each query image and gallery image for each person image in the query set, respectively. The performance of the ReID models is assessed using mean Average Precision (mAP) and Cumulative Matching Characteristics (CMC) at Rank1 and Rank5 matching accuracy. In the case of PRCC, CMC results are provided for our model under both single-shot and multi-shot settings.

\begin{figure*}[t]
	\centering
	\subfloat[Vanilla ResNet50 ]{\includegraphics[width=0.33\linewidth]{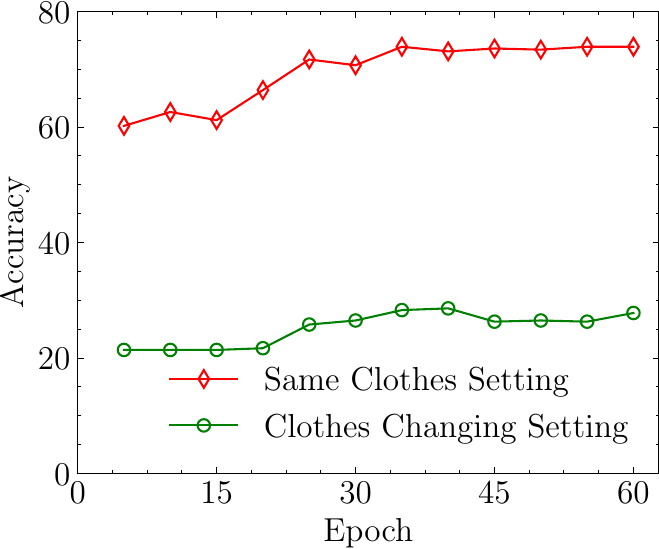}\label{fig:backbone}}
	\subfloat[Clothes Features $h_c$]{\includegraphics[width=0.33\linewidth]{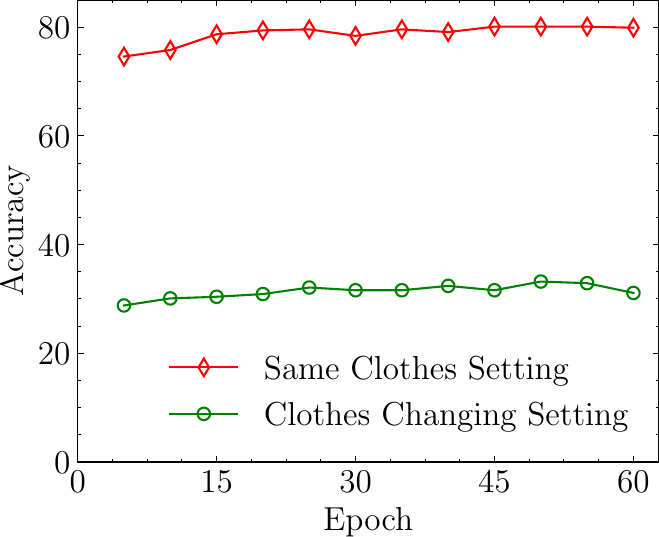}\label{fig:branch1}}
	\subfloat[Identity Features $h_{id}$]{\includegraphics[width=0.33\linewidth]{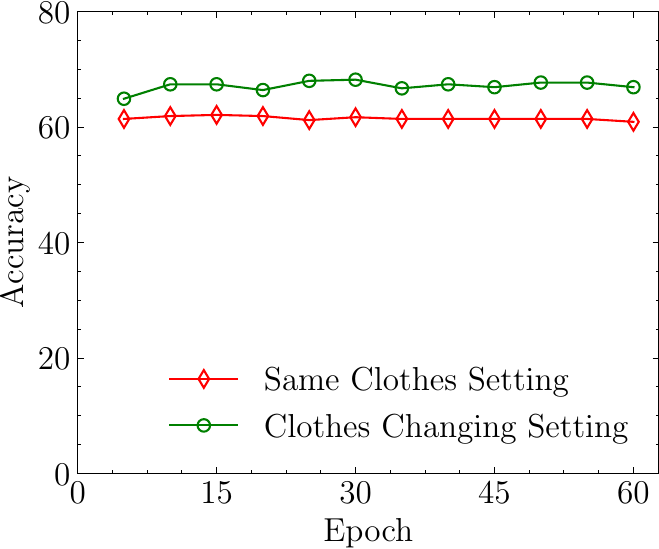}\label{fig:branch2}}
	\caption{ Comparing the effects of training ResNet50 with and without Diverse Norm on the LTCC dataset. }
	\label{fig:compare}
\end{figure*}

\noindent\textbf{Implementation Details}. We use ResNet50~\cite{he2016deep}  as the backbone architecture, omitting the final downsampling step for finer granularity. For LTCC and PRCC, following prior work \cite{gu2022clothes}, we apply global average pooling and global max pooling to the output feature map, concatenate the pooled features, and use task-specific BatchNorm~\cite{Ioffe2015BN} for normalization. Following \cite{Qian2020LTCC}, input images are resized to $384 \times 192$. Data augmentation includes random horizontal flipping, cropping, and erasing~\cite{zhong2020random}. The batch size is 64, with 8 individuals per batch, each represented by 8 images. We train the model with the Adam optimizer~\cite{Kingma2014Adam} for 60 epochs, using a channel attention module \cite{hu2018squeeze} as a projection layer before Diverse Norm. The initial learning rate is $3.5e^{-4}$, reduced by a factor of 10 every 20 epochs.

\subsection{Comparison with State-of-the-art Method}
We compared our proposed Diverse Norm with several state-of-the-art ReID models, both for standard ReID and those designed for clothes-changing scenarios. Standard ReID models include BoT \cite{luo2019bag}, PCB \cite{Sun2018Beyond}, MGN \cite{wang2018learning}, SCPNet \cite{fan2019scpnet}, ISGAN \cite{eom2019learning}, AIM \cite{yang2023good}, CAL \cite{gu2022clothes}, and OSNet \cite{zhou2019omni}, while clothes-changing ReID models include LTCC \cite{qian2020long}, CASE \cite{li2021learning}, FSAM \cite{Hong2021Finegrained}, CCFA \cite{han2023clothing} and 3DSL \cite{Chen2021Learning3D}. All models are implemented by Chan et al.\cite{chan2023learning}  using available codes, except for those without open-source code (such as \cite{li2021learning,qian2020long,wan2020person,yang2019person}), which were implemented based on paper descriptions.

\begin{figure}[t]
	\centering
	\includegraphics[width=1\linewidth]{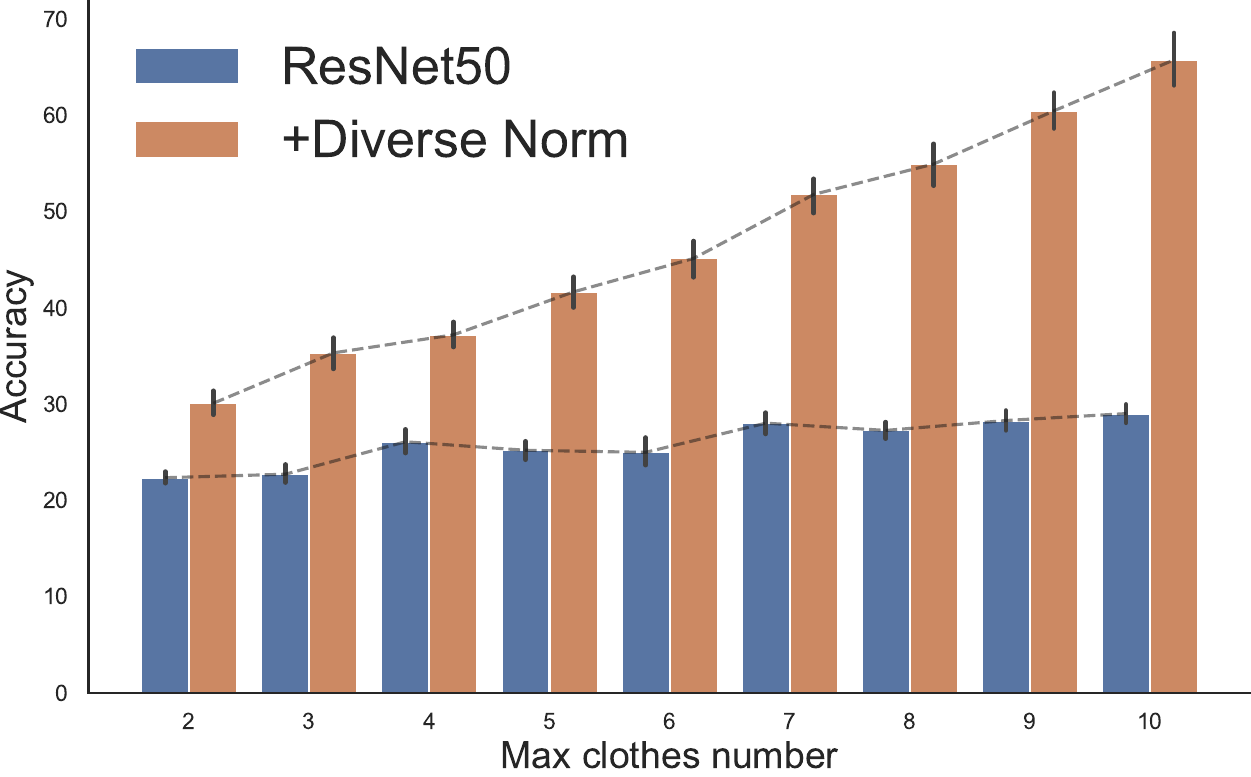}
	\caption{The connection between model performance and the number of clothes on LTCC.}
	\label{fig:dropclothes}
\end{figure}

\noindent\textbf{Evaluation on LTCC and PRCC.} The  results on clothes-changing ReID datasets, PRCC and LTCC, are presented in Table \ref{tab:res}. It's evident that all models experience a significant drop in accuracy when individuals change their clothes. 
In scenarios without clothing changes on PRCC, all models perform admirably, with most achieving over 96\% in Rank 1. Half of them even exceed 99\% in Rank 1, except ASE-SPT, which relies heavily on sketch images and achieves 72.09\%. Surprisingly, models designed for clothing changes do not outperform traditional models when clothing changes are absent. These specialized ReID models do not demonstrate a clear advantage over traditional models in the clothing-changing scenario in both datasets. Some traditional models even outperform clothing-changing models. For example, BoT achieves 44.92\% in Rank 1 on PRCC with clothing changes and 64.33\% and 28.12\% in the two settings on LTCC, outperforming most clothing-changing models. MGN and ISGAN even achieve over 50\% in Rank 1 on PRCC with clothing changes.

In Table \ref{tab:res}, several clothing-changing models, like CASE, 3APF, ReIDCaps, CAL, AIM, and AFD-Net, perform above average but fall short of achieving high results compared to Diverse Norm, especially in LTCC dataset. CASE encourages its identity encoder to overlook color information, potentially missing out on important clothing details other than color. Similarly, ReIDCaps learns clothes information across different dimensions of the output vector, making it challenging to ensure proper disentanglement of clothing features. AFD-Net also extracts and disentangles identity and clohtes features, where the latter is assumed to represent clothing features. Nevertheless, there is no assurance that identity details are successfully acquired and separated in the learning process. In contrast, the Diverse Norm method explicitly orthogonalize identity and clothing features, resulting in a more favorable optimization landscape. Surprisingly, by simply integrating this module into the backbone, we achieve state-of-the-art performance in both scenarios involving consistent clothing and clothing changes.

%\begin{table}
%\renewcommand\arraystretch{1.3}
%\caption{Comparing various query strategies.}\label{sum_spe}
%\begin{sc}A
%	\resizebox{0.48\textwidth}{!}{
% \begin{tabular}{l| c c c c| c c cc }
%	\hline
%	\multirow{3}*{Dataset}   &\multicolumn{4}{|c}{Seperately} &\multicolumn{4}{|c}{
%		Sum}\\
%	\cline{2-9}
%	&\multicolumn{2}{|c}{General/ SC}&\multicolumn{2}{|c}{
%		CC} &\multicolumn{2}{|c}{General/ SC}  &\multicolumn{2}{|c}{CC}\\
%	\cline{2-9}
%	\cline{2-9}&top-1 &mAP &top-1 &mAP  &top-1 &mAP &top-1 &mAP 
%	\\
%	\hline      LTCC  &\bf 78.9 & \bf 46.5  &\bf 63.3 & \bf  31.9 & 68.7 & 33.6 & 35.7 & 15.7  \\
%	PRCC    & \bf 100.0 &\bf 99.9  &\bf    58.9  & \bf 59.4 & 99.3 & 99.2  &    45.2 & 44.1 \\
%	VC-Clothes   &\bf  96.6  &\bf 95.8 &\bf 85.9 &\bf 83.2   & 94.4  &94.7  &72.8  &72.6   \\
%	LaST  & \bf 74.4  & \bf 29.7 & - & -  & 71.6  & 26.9 & - & -   \\
%	DeepChange  &  \bf  56.9  & \bf 19.4 & - & -   &    51.9 & 16.9  & - & - \\
%	\hline
%\end{tabular}}
%\end{sc}
%\end{table}
\begin{figure}[t]
	\centering
	\includegraphics[width=1\linewidth]{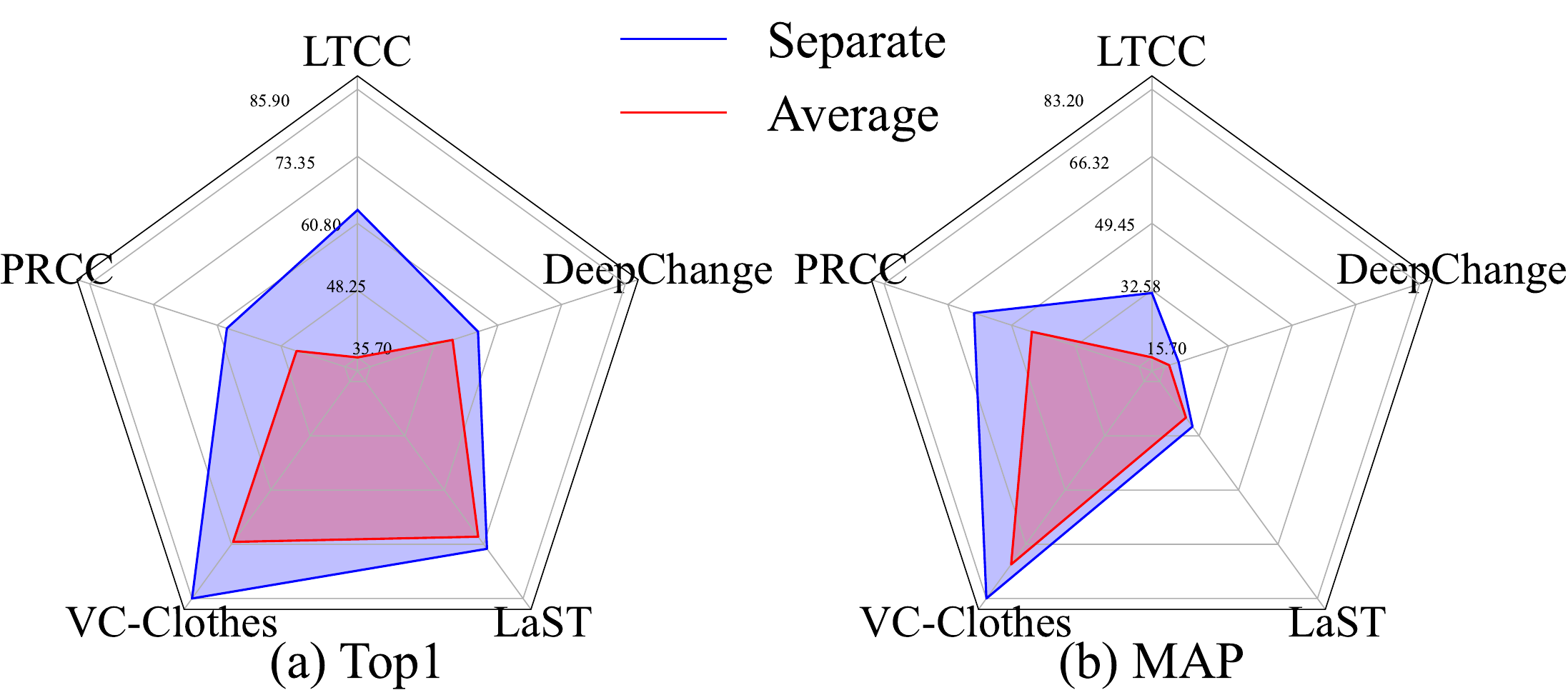}
	\caption{Comparing different query strategies.}
	\label{fig:top1}
\end{figure}

\noindent\textbf{Evaluation on VC-Clothes.}
To further demonstrate the superior of Diverse Norm,  the VC-Clothes dataset~\cite{Real28} is also considered, which is a synthetic virtual dataset generated using GTA5. It comprises a total of 19,060 images depicting 512 different identities captured from 4 distinct camera scenes. Each identity is seen wearing 1 to 3 different outfits, and it's worth noting that all images of the same identity captured by cameras 2 and 3 feature them wearing identical clothing. Consequently, many recent studies~\cite{Real28,Hong2021Finegrained} evaluate performance on a subset of this dataset obtained from cameras 2 and 3, referred to as the same-clothes (SC) setting. Additionally, results are reported for a subset from cameras 3 and 4, referred to as the clothes-changing (CC) setting. In this work, we adhere to these established evaluation settings for the sake of fair comparison. To assess the effectiveness, we benchmark it against two single-modality-based re-id methods MDLA~\cite{Qian2017MS} and PCB~\cite{Sun2018Beyond}, one clothes label based method CAL~\cite{gu2022clothes} as well as three multi-modality-based re-id methods Part-aligned~\cite{suh2018part}, FSAM~\cite{Hong2021Finegrained}, and 3DSL~\cite{Chen2021Learning3D} on the VC-Clothes dataset. The results are summarized in Table~\ref{tab:result2}.
Notably, with two expanding branches, Diverse Norm  consistently outperforms both the baseline and all the aforementioned state-of-the-art methods across all evaluation metrics, including general performance, the same-clothes setting, and the clothes-changing setting.

\noindent\textbf{Evaluation on Large-scale Datasets.}
LaST~\cite{LaST} and DeepChange~\cite{DeepChange} are two large-scale long-term person re-id datasets. LaST is a large-scale dataset with over 228,000 carefully labeled pedestrian images collected from movies, which is used to study the scenario where pedestrians have a large activity scope and time span.  DeepChange is a dataset of 178,000 bounding boxes from 1,100 person identities, collected over 12 months.
To ensure a fair comparison \cite{gu2022clothes}, we set the batch size to 64 and each batch contains 16 persons and 4 images per person. We report the performance of Diverse Norm and other methods: OSNet~\cite{OSNet}, ReIDCaps~\cite{huang2019beyond}, BoT~\cite{BoT},  mAPLoss~\cite{LaST}, and CAL \cite{gu2022clothes},  in a general setting. On DeepChange, we allow true matches to come from the same camera but different tracklets as query following \cite{DeepChange}.
As shown in \tableref{tab:result2}, Diverse Norm outperforms the baseline and state-of-the-art methods on both LaST and DeepChange. Notably, Diverse Norm remains effective on DeepChange when comparing with CAL, which uses the collection date as a proxy for clothes labels.

\subsection{Abalation Study}\label{src:abl}
%\noindent\textbf{Examining the Impact of Placing the Diverse Norm within ResNet-50.} The Diverse Norm introduced in this study can be seamlessly integrated at any position within the ResNet-50 backbone network. In our experiments, the ResNet-50 backbone consists of four distinct positions, denoted as ResBlock-$P_0$ to ResBlock-$P_3$. We conducted a systematic evaluation to assess the impact of integrating Diverse Norm at different positions on its performance.
%As illustrated in \tableref{tab:ablate}, our results demonstrate a gradual improvement in performance when Diverse Norm is integrated after ResBlock-$P_0$ to ResBlock-$P_2$. This indicates the need for a delicate balance between extracting general person-related features and task-specific features. Notably, the best results on both the PRCC and LTCC datasets were achieved when Diverse Norm was incorporated after position-2. However, introducing Diverse Norm after position-3 resulted in a slight performance decline. This can be attributed to inadequate learning of task-specific embeddings, making it challenging to distinguish between clothing-related and clothing-unrelated embeddings. 
%Based on this analysis, our default choice is to integrate Diverse Norm after position-2 of the backbone network, unless otherwise specified.  \\
\noindent\textbf{Examining the Impact of Clothes.} 
We observed an interesting phenomenon in \tableref{tab:res} where the improvement achieved by our method differs between the LTCC and PRCC datasets. We conjecture this discrepancy is due to the most significant difference between LTCC and PRCC: the number of clothes per person. In the PRCC dataset, each individual has only two outfits, whereas in the LTCC dataset, an individual can have up to 14 different outfits. We believe the number of outfits plays a crucial role in the performance of learning features that are unrelated to clothing, thereby affecting the overall performance of our method. To further evaluate our hypothesis, we conducted an experiment by removing different proportions of outfits in the LTCC dataset. The results are shown in \figref{fig:dropclothes}. We can observe that as the number of clothes decreases, the model's performance experiences a significant decline.

\noindent\textbf{Examining Different Query Strategies.} In this ablation study, we evaluated two distinct query strategies: one that computes Cosine similarity after summing the features, and another that calculates similarities separately before summation. The results are presented in  \figref{fig:top1}. Our findings reveal a substantial improvement when employing the separate calculation method.
We hypothesize that this improvement stems from the disproportionate representation of clothing in these images, which previously led to a skewed similarity metric that heavily favored clothing features. Due to the extreme imbalance in the proportion of clothing versus facial features within an image, even though the ResNet50 extracts facial features, they are often overshadowed by the dominant clothing features. By calculating Cosine  similarities separately, we mitigate this bias, allowing for a more balanced consideration of both clothing and identity-specific features.

\section{Conclusion}

In this paper, we are the first to identify the challenges faced by the CC-ReID baseline, specifically its difficulty in balancing the components of clothing and ID features. We introduced a novel technique called Diverse Norm for CC-ReID, which does not require any additional data, multi-modality inputs, or clothing labels. This method simply facilitates the separation of person features into an orthogonal space, effectively distinguishing between clothing and clothing-irrelevant attributes. Notably, Diverse Norm provides a straightforward and highly effective solution that can be seamlessly integrated into any CC-ReID dataset, surpassing current state-of-the-art methods that rely on additional information. We hope that Diverse Norm will become a commonly used module in future clothes-changing person re-identification methods.
\bibliography{aaai25}

\end{document}